# Building Energy Load Forecasting using Deep Neural Networks

Daniel L. Marino, Kasun Amarasinghe, Milos Manic
Department of Computer Science
Virginia Commonwealth University
Richmond, Virginia
marinodl@vcu.edu, amarasinghek@vcu.edu, misko@ieee.org

*Abstract*—Ensuring sustainability demands more efficient energy management with minimized energy wastage. Therefore, the power grid of the future should provide an unprecedented level of flexibility in energy management. To that end, intelligent decision making requires accurate predictions of future energy demand/load, both at aggregate and individual site level. Thus, energy load forecasting have received increased attention in the recent past, however has proven to be a difficult problem. This paper presents a novel energy load forecasting methodology based on Deep Neural Networks, specifically Long Short Term Memory (LSTM) algorithms. The presented work investigates two variants of the LSTM: 1) standard LSTM and 2) LSTM-based Sequence to Sequence (S2S) architecture. Both methods were implemented on a benchmark data set of electricity consumption data from one residential customer. Both architectures where trained and tested on one hour and one-minute time-step resolution datasets. Experimental results showed that the standard LSTM failed at one-minute resolution data while performing well in one-hour resolution data. It was shown that S2S architecture performed well on both datasets. Further, it was shown that the presented methods produced comparable results with the other deep learning methods for energy forecasting in literature.

## I. INTRODUCTION

Buildings are identified as a major energy consumer worldwide, accounting for 20%-40% of the total energy production [1]-[3]. In addition to being a major energy consumer, buildings are shown to account for a significant portion of energy wastage as well [4]. As energy wastage poses a threat to sustainability, making buildings energy efficient is extremely crucial. Therefore, in making building energy consumption more efficient, it is necessary to have accurate predictions of its future energy consumption.

At the grid level, to minimizing the energy wastage and making the power generation and distribution more efficient, the future of the power grid is moving to a new paradigm of smart grids [5], [6]. Smart grids are promising, unprecedented flexibility in energy generation and distribution [7]. In order to provide that flexibility, the power grid has to be able to dynamically adapt to the changes in demand and efficiently distribute the generated energy from the various sources such as renewables [8]. Therefore, intelligent control decisions should be made continuously at aggregate level as well as modular level in the grid. In achieving that goal and ensuring the reliability of the grid, the ability of forecasting the future demands is important. [6], [9].

Further, demand or load forecasting is crucial for mitigating uncertainties of the future [6]. In that, individual building level demand forecasting is crucial as well as forecasting aggregate loads. In terms of demand response, building level forecasting helps carry out demand response locally since the smart grids incorporate distributed energy generation [6]. The advent of smart meters have made the acquisition of energy consumption data at building and individual site level feasible. Thus data driven and statistical forecasting models are made possible [7].

Aggregate level and building level load forecasting can be viewed in three different categories: 1) Short-term 2) Medium-term and 3) Long-term [6]. It has been determined that the load forecasting is a hard problem and in that, individual building level load forecasting is even harder than aggregate load forecasting [6], [10]. Thus, it has received increased attention from researchers. In literature, two main methods can be found for performing energy load forecasting: 1) Physics principles based models and 2) Statistical and machine learning based models. Focus of the presented work is on the second category of statistical load forecasting. In [7], the authors used Artificial Neural Network (ANN) ensembles to perform the building level load forecasting. ANNs have been explored in detail for the purpose of all three categories of load forecasting [9], [11]-[13]. In [14], the authors use a support vector machines based regression model coupled with empirical mode decomposition to for long-term load forecasting. In [15], electricity demand is forecast using a kernel based multi-task learning methodologies. In [10], authors model individual household electricity loads using sparse coding to perform medium term load forecasting. In the interest of brevity, not all methods in literature are introduced in the paper. For surveys of different techniques used for load forecasting, readers are referred to [16], [17] and [8]. Despite the extensive research carried out in the area, individual site level load forecasting remains to be a difficult problem.

Therefore, the work presented in this paper investigates a deep learning based methodology for performing individual building level load forecasting. Deep learning allows models composed of multiple layers to learn representations in data. The use of multiple layers allow the learning process to be carried out with multiple layers of abstraction. A comprehensive overview and a review of deep learning methodologies can be found in [18]. In previous work for load forecasting using deep learning, authors of [6], explore Conditional Restricted Boltzmann Machines (CRBM) [19] and Factored Conditional Restricted Boltzmann Machines (FCBRM) [20] for building level load forecasting. Authors compare the two methods to several traditional methods including Support Vector Machines and Artificial Neural Networks. They conclude that the FCRBM method outperforms the other tested methodologies.

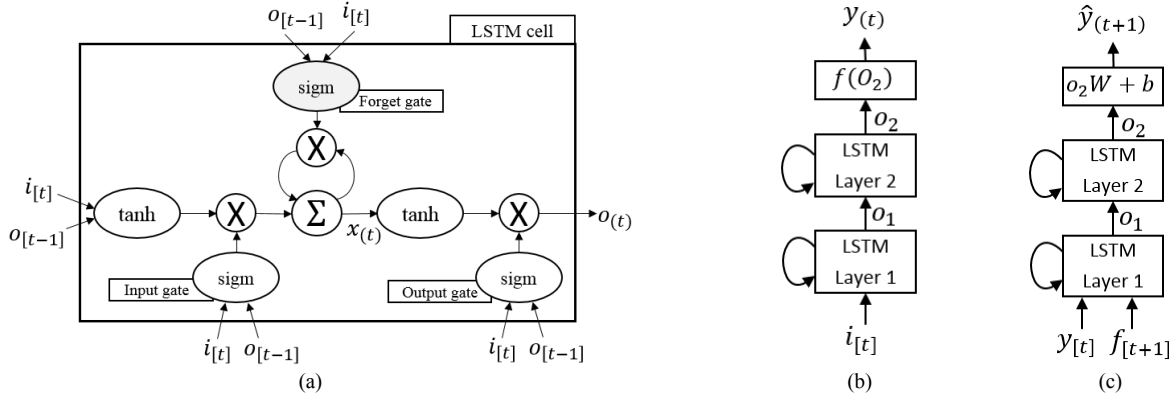

Fig. 1. (a) LSTM cell, (b) multilayer LSTM architecture, (c) architecture used for Load forecasting

In this work, the effectiveness of a different deep learning technique is explored for performing building level forecasting. The presented methodology uses Long Short Term Memory (LSTM) algorithm. Presented work investigates using two variations of the LSTM: 1) load forecasting using standard LSTM and 2) load forecasting using LSTM based Sequence to Sequence (S2S) architecture. Both methodologies are tested on a benchmark dataset which contained electricity consumption data for a single residential customer with time resolutions one minute and one hour. In order to compare the results, the same dataset used in [6] is used. Experimental results show that the LSTM based S2S architecture performs well on both types of datasets while the standard LSTM fails to perform well on the minute resolution data. Further, it's seen that the LSTM based algorithms manages to produce results comparable to the FCRBM and the CRBM in [6].

The rest of the paper is organized as follows. Section II provides background on the Long Short Term Memory algorithm. Section III elaborates the load forecasting using standard LSTM. Section IV elaborates load forecasting using LSTM based S2S architecture. Section V describes the dataset and the experimental results. Finally, Section VI concludes the paper.

## II. LONG SHORT TERM MEMORY

This section provides a brief introduction on the algorithm, Long Short Term Memory (LSTM).

Recurrent Neural Networks (RNN) are usually trained using either Back-propagation through time [21], or Real-Time Recurrent Learning [22] algorithm. Training with these methods often fails because of vanishing/exploding gradient. LSTM [23] is a recurrent neural-network that was specifically designed to overcome the problems of vanishing gradient, providing a model that is able to store information for long periods of time.

An LSTM network is comprised of memory cells with self-loops as shown in Fig. 1.a). The self-loop allows it to store temporal information encoded on the cell's state. The flow of information through the network is handled by writing, erasing and reading from the cell's memory state. These operations are handled by three gates respectively: 1) input gate, 2) forget gate and 3) output gate. The equations (1.a) through (1.f) express a single LSTM cell's operation.

$$i_g = \text{sigm}(i_{[t]}W_{ix} + o_{[t-1]}W_{im} + b_i) \quad (1.a)$$
$$f_g = \text{sigm}(i_{[t]}W_{fx} + o_{[t-1]}W_{fm} + b_f) \quad (1.b)$$
$$o_g = \text{sigm}(i_{[t]}W_{ox} + o_{[t-1]}W_{om} + b_o) \quad (1.c)$$
$$u = \tanh(i_{[t]}W_{ux} + o_{[t-1]}W_{um} + b_u) \quad (1.d)$$
$$x_{[t]} = f_g \circ x_{[t-1]} + i_g \circ u \quad (1.e)$$
$$o_{[t]} = o_g \circ \tanh(u) \quad (1.f)$$

where $i_g$ corresponds to the input gate, $f_g$ to the forget gate and $o_g$ to the output gate. $x_{[t]}$ is the value of the state at time step $t$, $o_{[t]}$ the output of the cell and $u$ is the update signal.

The input gate decides if the update signal should modify the memory state or not. This is done by using a sigmoid function as a "soft" switch, whose on/off state depends on the current input and previous output (Eq. 1.a). If the value of the input gate $(i_g)$ is close to zero, the update signal is multiplied by zero, therefore the state will not be affected by the update (Eq. 1.e). Forget and output gates work in a similar manner.

LSTM cells can be stacked in a multi-layer architecture to construct a network similar to the one shown in Fig. 1.b). The architecture shown in Fig. 1.b) is generally used to predict an outcome $\hat{y}_{[t]}$ at time $t \in \mathbb{N}$, given the set of all previous inputs $\{i_{[0]}, i_{[1]}, \dots, i_{[t]}\}$.

## III. LOAD FORECASTING USING DEEP NEURAL NETWORKS

This section elaborates the presented methodologies of Deep Neural Networks based load forecasting. The presented work investigates using two variants of the LSTM algorithm for load forecasting. This section first discusses the standard LSTM

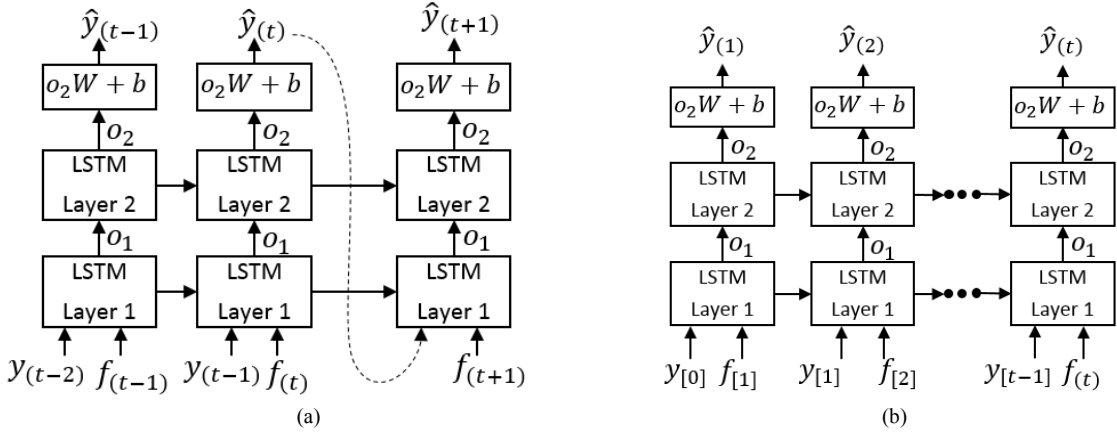

Fig. 2. (a) Use of LSTM network to make predictions in an arbitrary number of future time steps, (b) unrolled LSTM network for training through BPTT

based load forecasting methodology and then elaborates the LSTM based S2S architecture.

*A. Load Forecasting using Standard LSTM*

The objective of the presented methodology is to accurately estimate the electricity load (active power) for a time step or multiple time steps in the future, given historical electricity load data. I.e. having $M$ load measurements available, which can be expressed as:

$$y = \{y_{[0]}, y_{[1]}, \ldots, y_{[M-1]}\}, \quad (2)$$

where $y_{[t]}$ is the actual load measurement for time step $t$, the load for the following $T - M$ time steps should be predicted. The predicted load values can be expressed as:

$$\hat{y} = \{\hat{y}_{[M]}, \hat{y}_{[M+1]}, \ldots, \hat{y}_{[T]}\} \quad (3)$$

where $\hat{y}_{[t]}$ is the predicted load for time step $t$.

As the first technique, the standard LSTM algorithm was investigated. First model that was tested is illustrated in Fig. 1.c). The active power of the previous time step, and the date and time of the desired prediction are used as inputs for the model. The input vector can be expressed as:

$$\begin{aligned} i_{[t]} &= [y_{[t-1]} \quad day_{[t]} \quad day\_week_{[t]} \quad hour_{[t]}] \\ f_{[t]} &= [day_{[t]} \quad day\_week_{[t]} \quad hour_{[t]}] \end{aligned} \quad (4)$$

The output of the network, $\hat{y}_{[t]} \in \mathbb{R}$ is an estimation of the active power for the next time step. With this model, the electricity load for the next time step is predicted given a set of load measurements of the past. To predict further into the future, the predictions made by the model can be used as additional inputs for the next time step. Then, the input vector of the next time step can be expressed as:

$$i_{[t+1]} = [\hat{y}_{[t]} \quad day_{[t+1]} \quad day\_week_{[t+1]} \quad hour_{[t+1]}] \quad (5)$$

Fig. 2.a) illustrates this process.

To train the model, back-propagation through time(BPTT) is used. The network is unrolled by a fixed number of time steps as shown in Fig. 2.b). The resultant network can be seen as a very deep standard feedforward network with shared parameters. Therefore, standard backpropagation can be applied to train the network using a gradient based method such as Stochastic Gradient Descent (SGD).

The objective function that is minimized can be expressed as:

$$L = \sum_{t=1}^{M} (y_{[t]} - \hat{y}_{[t]})^2 \quad (6)$$

During the minimization process, a method called Norm clipping [24] is used to alleviate the exploding gradient problem.

For training, ADAM [25] algorithm is used as the gradient based optimizer, instead of SGD. ADAM outperformed SGD in terms of faster convergence and lower error ratios. The unrolling was implemented with 50 steps (*M*=50).

*B. Load forecasting using LSTM based sequence to sequence architecture*

In order to further improve the flexibility of the load forecasting methodology, a different architecture based on LSTM, called sequence to sequence (S2S), is explored. S2S is an architecture that was proposed to map sequences of different lengths [26].

Fig. 3 shows the S2S architecture that is employed for load forecasting. The architecture consists of two LSTM networks: encoder and a decoder. The task of the encoder is to convert input sequences of variable length and encode them in a fixed length vector, which is then used as the input state for the decoder. Then, the decoder generates an output sequence of length *n*. In this instance, that output sequence is the energy load forecast for the next *n* steps.

The main advantage of this architecture is that it allows inputs of arbitrary length. I.e. an arbitrary number of available load measurements of previous time steps can be used as inputs, to predict the load for an arbitrary number of future time steps.

To perform the prediction, $y$ (Eq. 2) is used as the input for the encoder together with the corresponding date and time for and time is used as input.

For training, the encoder network is pre-trained to minimize the following error:

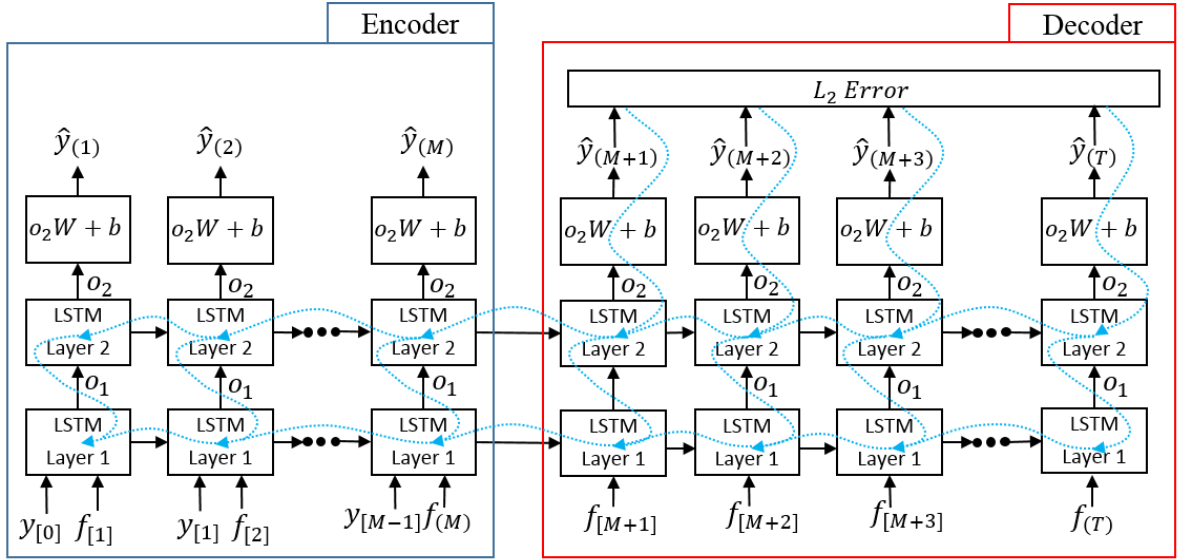

Fig. 3: S2S LSTM-based architecture for load forecasting (backpropagation signals shown in dashed arrows)

$$L_E = \sum_{t=1}^{M} (y_{[t]} - \hat{y}_{[t]})^2 \qquad (7)$$

Then the encoder plugged into the decoder network and train the two networks to reduce the objective function:

$$L_D = \sum_{t=M+1}^{T} (y_{[t]} - \hat{y}_{[t]})^2 \qquad (8)$$

Fig 3 also shows the path that back-propagation signals follow during training. Back-propagation signals are allowed to flow from the decoder to the encoder. Therefore, weights for both encoder and decoder are updated in order to minimize the objective function expressed in Eq. 8. Both decoder and encoder are updated because the pre-training of the encoder alone is insufficient to achieve good performance.

## IV. DATASET AND EXPERIMENTAL RESULTS

This section first introduces the dataset used for testing and then elaborates the experimental results obtained for the two models investigated.

### A. Dataset

The presented methods were implemented on a benchmark dataset of electricity consumption for a single residential customer, named "Individual household electric power consumption" [27]. The data set contained power consumption measurements gathered between December 2006 and November 2010 with one-minute resolution. The dataset contained aggregate active power load for the whole house and three sub metering for three sections for the house. In this paper, only the aggregate active load values for the whole house is used.

The dataset contained 2075259 measurements. Two typed of the data set were tested: 1) one-minute resolution data (original dataset) and 2) one-hour resolution data. The hourly resolution data were obtained by averaging the one-minute resolution data. Both architectures were tested on both one minute and one-hour resolution data. As in [6], the first three years where used to train the model and the last year was used as testing data. These ranges were chosen to be comparable with the work in [6].

### B. Experimental Results using standard LSTM

As the first experiment, predicting one step ahead with the standard LSTM network was attempted. It proved to be an easy task for the standard LSTM architecture, providing low error ratios with the test dataset. However, the model failed to provide an accurate forecast when the same model was used to predict further in the future, using the predictions as inputs, as mentioned in section III A (See Fig 2a).

Fig. 4 illustrates the performance of the model. For the first 60 hours, the actual load measurements on previous time steps were used as inputs to perform the prediction. Starting at hour 60, the predictions were introduced as inputs to generate a forecast for the next 60 hours. The figure shows how the model is incapable of providing accurate forecast for the last 60 hours, even with the forecast for one-time step ahead being very accurate.

It can be assumed that the reasons behind this behavior is that predicting the next step can be achieved with low error by

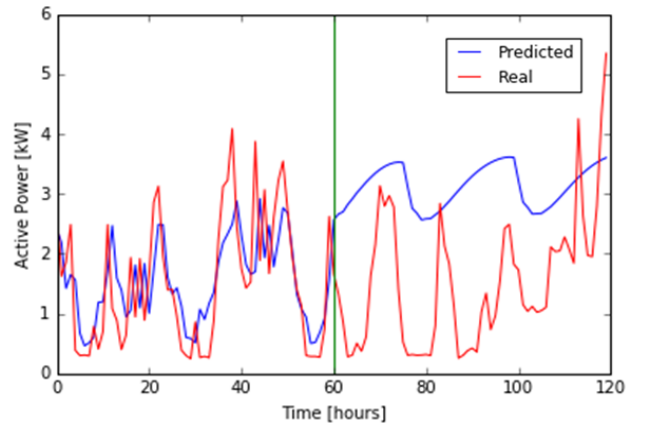

Fig 4: load forecasting for 60 hours using standard LSTM architecture

TABLE I. ERRORS FOR S2S ARCHITECTURE (ONE HOUR TIMESTEP, 60 HOURS FORECAST)

| Layers | Units | RMSE (Training) | RMSE (Testing) |
|---|---|---|---|
| 1 | 5 | 0.713 | 0.640 |
| 1 | 20 | 0.662 (0.677) | 0.657 (0.631) |
| 1 | 50 | 0.606 (0.697) | 0.686 (0.634) |
| 1 | 100 | 0.527 (0.697) | 0.729 (0.634) |
| 2 | 5 | 0.678 (0.688) | 0.642 (0.642) |
| 2 | 20 | 0.604 (0.7) | 0.675 (0.639) |
| 2 | 50 | 0.543 (0.689) | 0.727 (0.634) |
| 3 | 20 | 0.633 (0.696) | 0.665 (0.642) |

TABLE II. ERRORS FOR S2S ARCHITECTURE (SUMMARY)

|  | RMSE (Training) | RMSE (Testing) |
|---|---|---|
| 60 hours, one-hour resolution (2 Layers, 10 Units) | 0.701 | 0.625 |
| 60 min, one-minute resolution (2 Layers, 50 Units) | 0.742 | 0.667 |

simply bypassing the input from the current step straight to the output, this is because consequent measurements are very similar (when using one-minute resolution data). Therefore, if the network predicts that the load for the next time step is the same that the load on the current time. Thus, the neural network is learning a naïve mapping, where it generates an output equal to the input

Two approaches were investigated to solve the aforementioned problem. First was to introduce measurements from further in the past as inputs, for example 5 steps back, as opposed to inputting the load from the previous time step. This was done so that the input and output would different enough for the network to be able to learn a useful representation of the data. Fig. 5 shows the prediction made by the neural network after introducing the first 60 hours of load measurement and forecasting the next 60 hours. It can be seen that the architecture provides an estimation that follows the general trend of the future load. This method produced accurate results when used with hourly data, but failed to perform well with one-minute resolution data.

Given that a delay of the input was not sufficient to provide a useful model for one-minute time steps, the second approach that was tested was to experiment with a S2S, LSTM-based architecture. The results of this approach are elaborated in the next subsection.

### C. Experimental Results using LSTM based S2S architecture

As mentioned, LSTM-based S2S architecture was proposed to alleviate the problems encountered in the previous section.

Fig. 3 illustrates how the available load measurements are only introduced on the encoder, while the decoder's inputs are only date and time. This architecture allows us to prevent the decoder from learning the naïve mapping of passing the input straight to the output as explained in the previous section.

Table I shows the Root Mean Square Error (RMSE) on training and testing datasets for different number of layers and units using the S2S architecture for data with one hour resolution. The table shows the value of the errors at the end of the training and in parenthesis the lowest error obtained on testing dataset that was found during training.

It can be seen that the proposed architecture is able to produce very low errors on training dataset. Further, it was noticed that increasing the capacity of the network by increasing the number of layers and units only improves error on training dataset. Fig. 6 shows an example of how well the model performs on training dataset using a 2 layer network with 50 units in each layer. However, increasing the capacity of the network did not improve performance on testing data. In order to improve accuracy on testing data Dropout [28] was used as regularization methodology. Table II shows the errors obtained using Dropout for the one-minute and for the one-hour datasets. The predictions were made for 60 time steps in the future. The results shown on Table II are comparable to the results shown in [6] by using FCRBM. Fig. 7 shows an example of the prediction on the testing dataset.

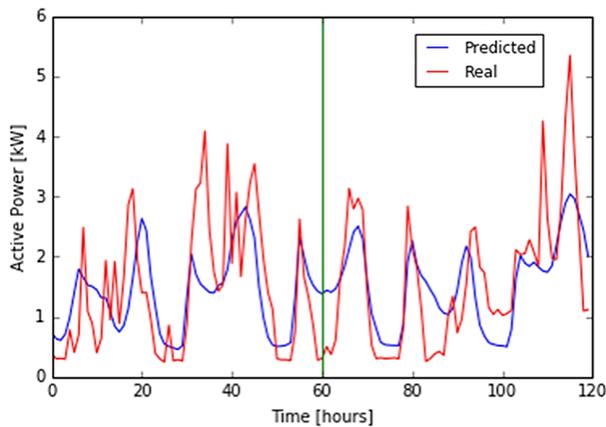

Fig 5: load forecast for 60 hours using standard LSTM and delayed input

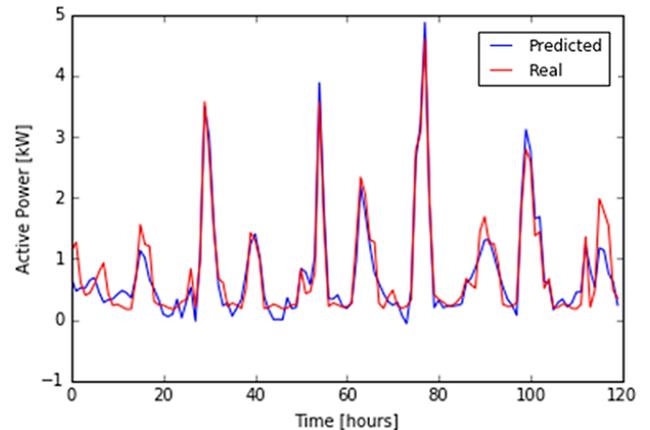

Fig 6: Prediction results for the training dataset using S2S model

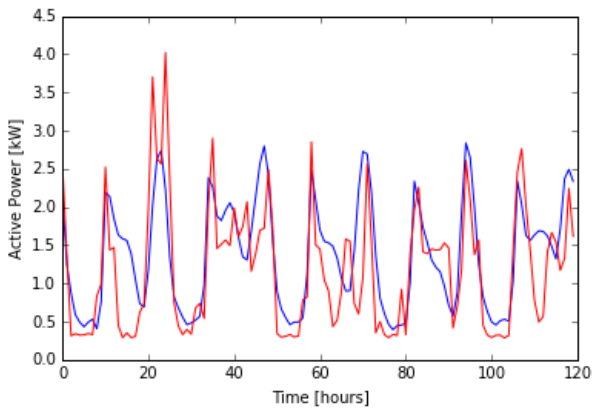

Fig 7: Prediction results for the testing dataset using S2S model

V. CONCLUSIONS

The goal of the presented work was to investigate the effectiveness in using LSTM based neural networks for building level energy load forecasting. This paper presented two LSTM based neural networks architectures for load forecasting. Both were trained and tested for one hour and one minute time-step resolution data. The standard LSTM architecture was unable to accurately forecast loads using one-minute resolution, the S2S LSTM-based architecture performed well in both datasets. Further, the S2S architecture provides a flexible model that is able to receive an arbitrary number of previous available load measurements as input to estimate the load for an arbitrary number of future time steps. The presented S2S model was able to produce comparable results to the FCRBM based results presented in [6] for the same dataset. However, to compare the effectiveness of these algorithms, both algorithms need to be tested on different real world datasets as future work. Further, we plan on investigating other deep learning algorithms as well as other regularization approaches to improve the generalization of the models.